\documentclass[10pt,twocolumn,letterpaper]{article}

\usepackage{cvpr}              % To produce the CAMERA-READY version
\usepackage[nohyperlinks, printonlyused, nolist]{acronym}
\usepackage[accsupp]{axessibility}

\definecolor{cvprblue}{rgb}{0.21,0.49,0.74}
\usepackage[pagebackref,breaklinks,colorlinks,allcolors=cvprblue]{hyperref}
\usepackage{multirow}
\usepackage{siunitx}
\usepackage{pifont}
\newcommand{\tmark}{\ding{51}}%
\def\paperID{*****} % *** Enter the Paper ID here
\def\confName{CVPR}
\def\confYear{2025}

\title{SLRTP2025 Sign Language Production Challenge: Methodology, Results, and Future Work}

\author{
Harry Walsh\textsuperscript{1*}, Ed Fish\textsuperscript{1*}, Ozge Mercanoglu Sincan\textsuperscript{1*}, Mohamed Ilyes Lakhal\textsuperscript{1*}, Richard Bowden\textsuperscript{1*}, \\ Neil Fox\textsuperscript{2}, Bencie Woll\textsuperscript{3}, Kepeng Wu\textsuperscript{4}, Zecheng Li\textsuperscript{4}, Weichao Zhao\textsuperscript{4}, Haodong Wang\textsuperscript{4}, \\ Wengang Zhou\textsuperscript{4}, Houqiang Li\textsuperscript{4}, Shengeng Tang\textsuperscript{5}, Jiayi He\textsuperscript{5}, Xu Wang\textsuperscript{5}, Ruobei Zhang\textsuperscript{5}, \\ Yaxiong Wang\textsuperscript{5}, Lechao Cheng\textsuperscript{5}, Meryem Tasyurek\textsuperscript{6}, Tugce Kiziltepe\textsuperscript{6}, Hacer Yalim Keles\textsuperscript{6} \\ \textsuperscript{1}University of Surrey, \textsuperscript{2}University of Birmingham, \textsuperscript{3}University College London, \\ \textsuperscript{4}University of Science and Technology of China, \textsuperscript{5}Hefei University of Technology, \textsuperscript{6}Hacettepe University \\ {\tt\small \{harry.walsh, edward.fish, o.mercanoglusincan, m.lakhal, r.bowden\}@surrey.ac.uk} \\ {\small \textsuperscript{*} denotes key workshop and challenge organisers.}
}

\begin{document}
 
% \include{symbols}
%USAGE:
%
%\ac{hog}
%force long = \acl{hog}
%force short = \acs{hog}
%
%
% \section{Nomenclature}
\markboth{Nomenclature}{Nomenclature}
\begin{acronym}[iccv] %replace parameter with longest acronym in list

%Conferences
\acro{bmvc}[BMVC]{British Machine Vision Conference}
\acro{iccv}[ICCV]{International Conference on Computer Vision}

%Misc
\acro{ai}[AI]{Artificial Intelligence}
\acro{ar}[AR]{Augmented Reality}
\acro{sdk}[SDK]{Software Development Kit}

\acrodefplural{rnn}[RNNs]{Recurrent Neural Networks}
\acrodefplural{cnn}[CNNs]{Convolutional Neural Networks}
\acrodefplural{hmm}[HMMs]{Hidden Markov Models}
\acrodefplural{gru}[GRUs]{Gated Recurrent Units}
\acrodefplural{crf}[CRFs]{Conditional Random Fields}
\acrodefplural{gan}[GANs]{Generative Adversarial Networks}
\acrodefplural{gpu}[GPUs]{Graphic Processing Units}

\acrodefplural{mdn}[MDNs]{Mixture Density Networks}
%acronym list

% A
\acro{asl}[ASL]{American Sign Language}
% B
\acro{btg}[BTG]{Bracketing Transduction Grammar}
\acro{bpe}[BPE]{Byte Pair Encoding}
\acro{bsl}[BSL]{British Sign Language}
\acro{bleu}[BLEU]{Bilingual Evaluation Understudy}
\acro{bobsl}[BOBSL]{BBC-Oxford British Sign Language}
\acro{blstm}[BLSTM]{Bidirectional Long Short-Term Memory}
\acro{bslcpt}[BSLCP\textbf{T}]{BSL Corpus \textbf{T}}
% C 
\acro{cnn}[CNN]{Convolutional Neural Network}
\acro{crf}[CRF]{Conditional Random Field}
\acro{cslr}[CSLR]{Continuous Sign Language Recognition}
\acro{ctc}[CTC]{Connectionist Temporal Classification}
\acro{c4a}[C4A]{Content4All}
% D
\acro{dl}[DL]{Deep Learning}
\acro{dgs}[DGS]{German Sign Language - Deutsche Gebärdensprache}
\acro{dsgs}[DSGS]{Swiss German Sign Language - Deutschschweizer Geb\"ardensprache}
\acro{dtw}[DTW]{Dynamic Time Warping}
\acro{dtwmje}[DTW-MJE]{Dynamic Time Warping Mean Joint Error}
% F
\acro{fc}[FC]{Fully Connected}
\acro{ff}[FF]{Feed Forward}
\acro{fps}[fps]{frames per second}
% G
\acro{gan}[GAN]{Generative Adversarial Network}
\acro{gpu}[GPU]{Graphics Processing Unit}
\acro{gru}[GRU]{Gated Recurrent Unit}
\acro{gtpt}[G2PT]{Gloss-to-Pose Transformer}
\acro{gtp}[G2P]{Gloss-to-Pose}
\acro{gts}[G2S]{Gloss-to-Sign}
\acro{gtt}[G2T]{Gloss-to-Text}
\acro{gs}[GS]{Gloss Selection}
\acro{gr}[GR]{Gloss Reordering}
\acro{gt}[GT]{ground truth}
% H
\acro{hmm}[HMM]{Hidden Markov Model}
\acro{hpe}[HPE]{Hand Pose Enhancer}
\acro{hoh}[HOH]{Hard of Hearing}
\acro{hns}[HamNoSys]{Hamburg Notation System}

% I
\acro{isl}[ISL]{Irish Sign Language}
% L
\acro{lstm}[LSTM]{Long Short-Term Memory}
\acro{lsf}[LSF]{French Sign Language}
% M
\acro{mha}[MHA]{Multi-Headed Attention}
\acro{mo}[MoCap]{Motion Capture}
\acro{mtc}[MTC]{Monocular Total Capture}
\acro{mse}[MSE]{Mean Squared Error}
\acro{mdn}[MDN]{Mixture Density Network}
\acro{mdgs}[MeineDGS]{Meine DGS Annotated}
\acro{mdgst}[mDGS\textbf{T}]{meineDGS\textbf{T}}
\acro{mdgsth}[mDGS\textbf{T}-\textbf{H}]{meineDGS\textbf{T}-\textbf{HARD}}
\acro{mdgste}[mDGS\textbf{T}-\textbf{E}]{meineDGS\textbf{T}-\textbf{EASY}}
\acro{mt}[MT]{Machine Translation}

% N
\acro{nmt}[NMT]{Neural Machine Translation}
\acro{nlp}[NLP]{Natural Language Processing}
\acro{nar}[NAR]{Non-AutoRegressive}
\acro{nsvq}[NSVQ]{Noise Substitution Vector Quantization}
% P
\acro{ph12}[PHOENIX12]{RWTH-PHOENIX-Weather-2012}
\acro{ph14}[PHOENIX14]{RWTH-PHOENIX-Weather-2014}
\acro{ph14t}[PHOENIX14\textbf{T}]{RWTH-PHOENIX-Weather-2014\textbf{T}}
\acro{pof}[POF]{Part Orientation Field}
\acro{pos}[POS]{Part Of Speech}
\acro{pt}[PT]{Progressive Transformer}
\acro{paf}[PAF]{Part Affinity Field}
\acro{pttt}[P2TT]{Pose-to-Text Transformer}
\acro{ptt}[P2T]{Pose-to-Text}
\acro{pts}[P2S]{Pose-to-Sign}
\acro{ptgtt}[P2G2T]{Pose-to-Gloss-to-Text}
\acro{pca}[PCA]{Principal Component Analysis}
% R
\acro{relu}[RELU]{Rectified Linear Units}
\acro{rnn}[RNN]{Recurrent Neural Network}
\acro{rouge}[ROUGE]{Recall-Oriented Understudy for Gisting Evaluation}
% V
\acro{vqvae}[VQ-VAE]{Vector Quantized Variational Autoencoders}
\acro{vq}[VQ]{Vector Quantisation}
\acro{vae}[VAE]{Variational Autoencoders}
% S
\acro{sgd}[SGD]{Stochastic Gradient Descent}
\acro{sla}[SLA]{Sign Language Assessment}
\acro{slr}[SLR]{Sign Language Recognition}
\acro{slt}[SLT]{Sign Language Translation}
\acro{slp}[SLP]{Sign Language Production}
\acro{smt}[SMT]{Statistical Machine Translation}
\acro{slo}[SLO]{Spoken Language Order}
\acro{so}[SO]{Sign Language Order}
\acro{snr}[S\&R]{Select and Reorder}
\acro{sse}[SSE]{Sign Supported English}
\acro{stt}[S2T]{Sign-to-Text}
\acro{stgtt}[S2G2T]{Sign-to-Gloss-to-Text}
\acro{sio}[SIO]{Sign Language Order}
\acro{spo}[SPO]{Spoken Language Order}
% T
\acro{ttgt}[T2GT]{Text-to-Gloss Transformer}
\acro{ttpt}[T2PT]{Text-to-Pose Transformer}
\acro{ttp}[T2P]{Text-to-Pose}
\acro{ttg}[T2G]{Text-to-Gloss}
\acro{ttg++}[T2G++]{Text-to-Gloss++}
\acro{tth}[T2H]{Text-to-HamNoSys}
\acro{ttgth}[T2G2H]{Text-to-Gloss-to-HamNoSys}
\acro{ttgtp}[T2G2P]{Text-to-Gloss-to-Pose}
\acro{tts}[T2S]{Text-to-Sign}
\acro{ttsse}[T2SSE]{Text to Sign Supported English}
\acro{ttspog}[T2SPOG]{Text to Spoken Language Order Gloss}
% W
\acro{wer}[WER]{Word Error Rate}
\acro{wmt14}[WMT2014]{WMT2014 German-English}

\end{acronym}

\maketitle
\begin{abstract}
\label{sec:abstract}

\noindent
Sign Language Production (SLP) is the task of generating sign language video from spoken language inputs. The field has seen a range of innovations over the last few years, with the introduction of deep learning-based approaches providing significant improvements in the realism and naturalness of generated outputs. However, the lack of standardized evaluation metrics for SLP approaches hampers meaningful comparisons across different systems. To address this, we introduce the first Sign Language Production Challenge, held as part of the third SLRTP Workshop at CVPR 2025. The competition's aims are to evaluate architectures that translate from spoken language sentences to a sequence of skeleton poses, known as Text-to-Pose (T2P) translation, over a range of metrics. For our evaluation data, we use the RWTH-PHOENIX-Weather-2014\textbf{T} dataset, a German Sign Language - Deutsche Gebärdensprache (DGS) weather broadcast dataset. In addition, we curate a custom hidden test set from a similar domain of discourse.

    This paper presents the challenge design and the winning methodologies.  The challenge attracted 33 participants who submitted 231 solutions, with the top-performing team achieving BLEU-1 scores of 31.40 and DTW-MJE of 0.0574. The winning approach utilized a retrieval-based framework and a pre-trained language model.  As part of the workshop, we release a standardized evaluation network, including high-quality skeleton extraction-based keypoints establishing a consistent baseline for the SLP field, which will enable future researchers to compare their work against a broader range of methods.
\end{abstract}
    
\section{Introduction}
\label{sec:intro}

Sign languages, like spoken languages, are complex systems with distinct grammar and vocabularies. They are independent and fully fledged languages with a unique structure, using manual and non-manual features asynchronously to convey information \cite{sutton1999linguistics}. Manual features can be defined as the physical motion, location, and shape of the hands and arms, while non-manual features include facial expressions, head movements, and body posture. Translating between spoken and signed languages presents a significant communication challenge, typically requiring expert human interpreters rather than simple word-to-sign substitution. However, the scarcity of interpreters limits access to information for the Deaf community. Sign Language Production (SLP), the task of generating sign language from spoken language inputs, has the potential to be part of a solution to improving accessibility. 

Over the last three decades, research into SLP has made significant progress \cite{tamura1988recognition}. While early approaches utilized graphical avatars and rule-based translation systems \cite{bangham2000virtual, cox2002tessa, efthimiou2012dicta, elghoul2011websign, zwitserlood2004synthetic}, these architectures often produced robotic and unnatural movement. However, more recent progress in deep learning-based methods has significantly enhanced the realism and naturalness of sign language generation \cite{saunders2020progressive, saunders2021signing, walsh2024data, walsh2024sign, CHEN2024104050, xie2022vector}. Approaches to SLP commonly decompose the task into several stages, using intermediate representations such as linguistic annotation \cite{stoll2020text2sign, walsh2022changing, walsh2024select}, skeleton pose \cite{walsh2024sign, saunders2020progressive}, and parametric human models \cite{stoll2022there, zuo2024simple, baltatzis2024neural}.

Despite these advancements, progress in the field has been impeded by the absence of standardized evaluation metrics, which hinders meaningful comparisons across SLP systems. To address this issue, and as part of the SLRTP Workshop\footnote{Workshop website: \url{https://slrtpworkshop.github.io/}}, a sign language production challenge was held, attracting 33 participants who submitted 231 solutions\footnote{Challenge website: \url{https://www.codabench.org/competitions/4854/}}. In this manuscript, we present the approaches from the top three performing teams. The competition focused on developing robust systems for spoken language to sign language translation. As part of the evaluation pipeline, we propose a novel metric, total distance, to help score the expressiveness of the productions. Given previous work has noted issues with regression to the mean, the proposed  metric provides an improved qualitative measure for translation quality. We release the evaluation pipeline\footnote{Challenge Evaluation Code: \url{https://github.com/walsharry/SLRTP-Sign-Production-Evaluation/tree/main}}, with the hope that this helps establish a consistent baseline for SLP work, enabling future researchers to conduct comprehensive comparisons with a wider range of methods. 

The rest of this paper is organized as follows. Section \ref{sec:related_work} reviews related work in the field of \ac{slp}. Sections \ref{sec:design} and \ref{sec:dataset} describe the design and dataset used in the challenge. Section \ref{sec:protocol} details the evaluation protocol. Section \ref{sec:methodology} outlines the methodology used by the top three teams, followed by Section \ref{sec:results}, which presents the results. Finally, Section \ref{sec:conclusion} concludes the paper and suggests directions for future work.

\section{Related Work}
\label{sec:related_work}

\subsection{Sign Language Production}

The shortage of qualified sign language interpreters has motivated the development of \ac{slp} systems. Computational sign language research started in the early 90s \cite{tamura1988recognition}, with the first approaches being avatar-based \cite{bangham2000virtual, cox2002tessa, efthimiou2012dicta, elghoul2011websign, zwitserlood2004synthetic}. However, the use of poor quality avatars has consistently received criticism from the Deaf community due to their unrealistic appearance and robotic motion, which limits comprehension \cite{glauert2006vanessa}. Some of these systems looked for ``legal" phrases in the input text and then mapped them to pre-defined sign language animations \cite{cox2002tessa}. Another approach uses \ac{mo} to incorporate more fluid motion dynamics, but this requires specialised capture equipment which then limits the vocabulary size \cite{gibet2016interactive}.

The field then progressed to Statistical Machine Translation (SMT), allowing for automatic rule learning and improved generalizability \cite{Bungeroth2004StatisticalSL, kanis2006czech, othman2011statisticalsignlanguagemachine}. These models were able to use the context of the input text to solve for ambiguity. However, such approaches require handcrafted features to be extracted before being used in an ensemble of models. 

The introduction of deep learning has resulted in more data-driven approaches, that can learn the mapping between spoken and sign languages. The first deep \ac{slp} pipeline broke down the task into three steps: first, a translation from \ac{ttg}, followed by a \ac{gtp} look-up, and finally a \ac{pts} video generation step \cite{stoll2018sign}. However, this approach was unable to blend the motion between signs, and due to the low-resolution output key features such as facial expressions were lost.

Later approaches attempted to synthesize poses from textual input. Zelinka et al. \cite{9093516} utilized a \ac{rnn} to predict a 7-frame pose sequence for each word in the sentence. As a result, the length of each prediction is dependent on the number of words in the sentence. This limited the realism of the outputs. After, Saunders et al. \cite{saunders2020progressive} introduced the progressive transformer, the first to learn a direct mapping between text and pose using a transformer architecture. This approach was able to generate more realistic outputs and deal with the variable length nature of sign language. Further improvements were achieved by adding adversarial training and a \ac{mdn} \cite{saunders2020adversarial, saunders2021mixed}. A non-autoregressive transformer was also introduced to improve the speed of the model \cite{huang2021towards}. Recently, diffusion has been applied to the tasks \cite{CHEN2024104050, 10.1145/3663572} and has shown improvements, but this work relies on linguistic annotation to guide the process. Similar to most methods discussed so far, large performance increases can be gained by introducing gloss\footnote{Gloss is the written words associated with a sign.} annotations \cite{saunders2020progressive, huang2021towards, saunders2020adversarial, saunders2021mixed}. However, using gloss is a major limiting factor when scaling to larger domains of discourse. 

Models that attempted to directly regress a pose sequence from text input often struggled with regression to the mean. This leads to the generation of less expressive outputs. To address this, some works have applied \ac{vq} to the task, where the model is used to learn a set of discrete codes that map to a small sequence of poses. This then serves as the lexicon for translation. Some of which predict the codes from text \cite{hwang2023autoregressive, walsh2024data}, while others start from glosses \cite{xie2022vector}.  

Instead of learning units, others have used a dictionary of pre-recorded signs. They are guaranteed to be expressive and therefore produce comprehensible signing. Simply concatenating the signs can create an unnatural sequence, so Walsh et al. \cite{walsh2024sign} used a 7-step pipeline to create smooth transitions, while others employed a transformer \cite{zuo2024simple} or a diffusion model \cite{tang2024discrete} for the task. Alternatively, Saunders et al. \cite{saunders2022signing} proposed a novel keyframe selection network to learn the co-articulate between signs.

While the works discussed so far have employed skeleton poses as a representation for sign language the source can vary. Some researchers utilize MediaPipe \cite{lugaresi2019mediapipe}, whereas others opt for OpenPose \cite{cao2019openposerealtimemultiperson2d}. Significant variation exists across methods, with different authors using a different subset of keypoints. This has hindered meaningful comparison between these approaches. 

Beyond skeleton pose, human parametric models have also been used for the \ac{slp} task \cite{stoll2022there, zuo2024simple, baltatzis2024neural}. All of which leverage SMPL-X \cite{loper2015smpl}, a human model with parameters to describe the face, hands, and body. Other methods have directly predicted RGB video frames \cite{guo2024unsupervisedsignlanguagetranslation, xie2023signlanguageproductionlatent}

A substantial body of research exists; however, direct comparison between these works is challenging due to the heterogeneity of output representations. We observe significant variability, particularly in skeleton pose representations. Furthermore, extraction and normalization techniques diverge considerably across studies. This inconsistency impedes the ability to discern whether performance improvements stem from novel algorithmic contributions or advancements in skeleton extraction and normalization methodologies. This issue is compounded by the limited public availability of codebases, which renders direct comparison between state-of-the-art approaches exceedingly difficult and highlights the need for more standardized evaluation metrics for \ac{slp}.
\section{Challenge Design}
\label{sec:design}

The objective of this challenge was to generate continuous sign language sequences from spoken language inputs. We utilized a skeleton pose representation for sign language. As seen in the literature, this serves as a common intermediate representation and has been shown to be capable of driving photo-realistic signer generation \cite{saunders2020everybody, pelykh2024givinghanddiffusionmodels, fang2023signdifflearningdiffusionmodels}.

The challenge comprised two phases. Initially, during the development phase, teams were provided with train and dev splits from the \ac{ph14t} dataset \cite{camgoz2018neural}.  Throughout this phase, participants could submit solutions to the test set via the online platform, receiving evaluation scores for their skeleton pose predictions. Subsequently, in the final week, the competition transitioned to the test phase. Participants were presented with 500 spoken language sentences from the hidden test set and tasked with submitting their model's predictions. The competition was held on the Codabench platform \cite{Xu_2022}, an open-source framework for running competitions, enabling the automated evaluation of code and results. The competition spanned 49 days, starting on January 13, and finishing on March 3, 2025. During this time, 33 participants contributed 231 solutions. During the development phase, participants were limited to 100 submissions per day, or a total of 3,000 submissions throughout the competition. During the final test phase, participants were limited to just three submissions. This constraint was implemented to minimize the likelihood of participants overfitting to specific data portions or employing random initializations to achieve performance gains.

In the final stage, six teams submitted solutions that outperformed the baseline approach. The top-performing teams were requested to submit a fact sheet detailing their approach. The teams with the highest scores, and who shared information about their code and methodology, were selected as the winners. 

For the final ranking, in \cref{sec:results}, we employ Pareto dominance. A solution dominates another if it performs at least as well in all metrics and demonstrably better in at least one. This method ensures that no single metric is arbitrarily favoured, providing a balanced and fair evaluation.

\section{Challenge Datasets}
\label{sec:dataset}

For the SLRTP 2025 CVPR challenge, we use the RWTH-PHOENIX-Weather-2014T (\ac{ph14t}) dataset \cite{camgoz2018neural} and a custom hidden test set. Here we provide details on each.

\subsection{PHOENIX14\textbf{T}}

The \ac{ph14t} dataset is one of the most widely used benchmarks for research in sign language recognition and translation. It contains continuous sign language videos along with their corresponding gloss annotations and spoken language subtitles. The dataset is derived from weather forecast broadcasts in German Sign Language (DGS).

The dataset is split into training, development, and test sets. The training set contains 7096 videos, the development set contains 519 videos, and the test set contains 642 videos.

\subsection{Hidden Test Set}

The hidden test set for this challenge was sourced from the Phoenix broadcast channel, consistent with the original \ac{ph14t} dataset. This data was collected as part of the EASIER project\footnote{https://www.project-easier.eu/}. However, unlike the original dataset, it contains minimal manual annotation. Therefore, we first searched the subtitles for sections most relevant to weather-related discussions, to maintain the original domain of discourse. We then cropped these sections from the original videos. To ensure that the selected segments were indeed from weather broadcasts, we manually verified a subset of the total videos. Finally, we randomly selected 500 sentences to form the hidden test set.

\subsection{Skeleton Representation}

\begin{figure}[htbp]
    \centering
    \includegraphics[width=0.45\textwidth]{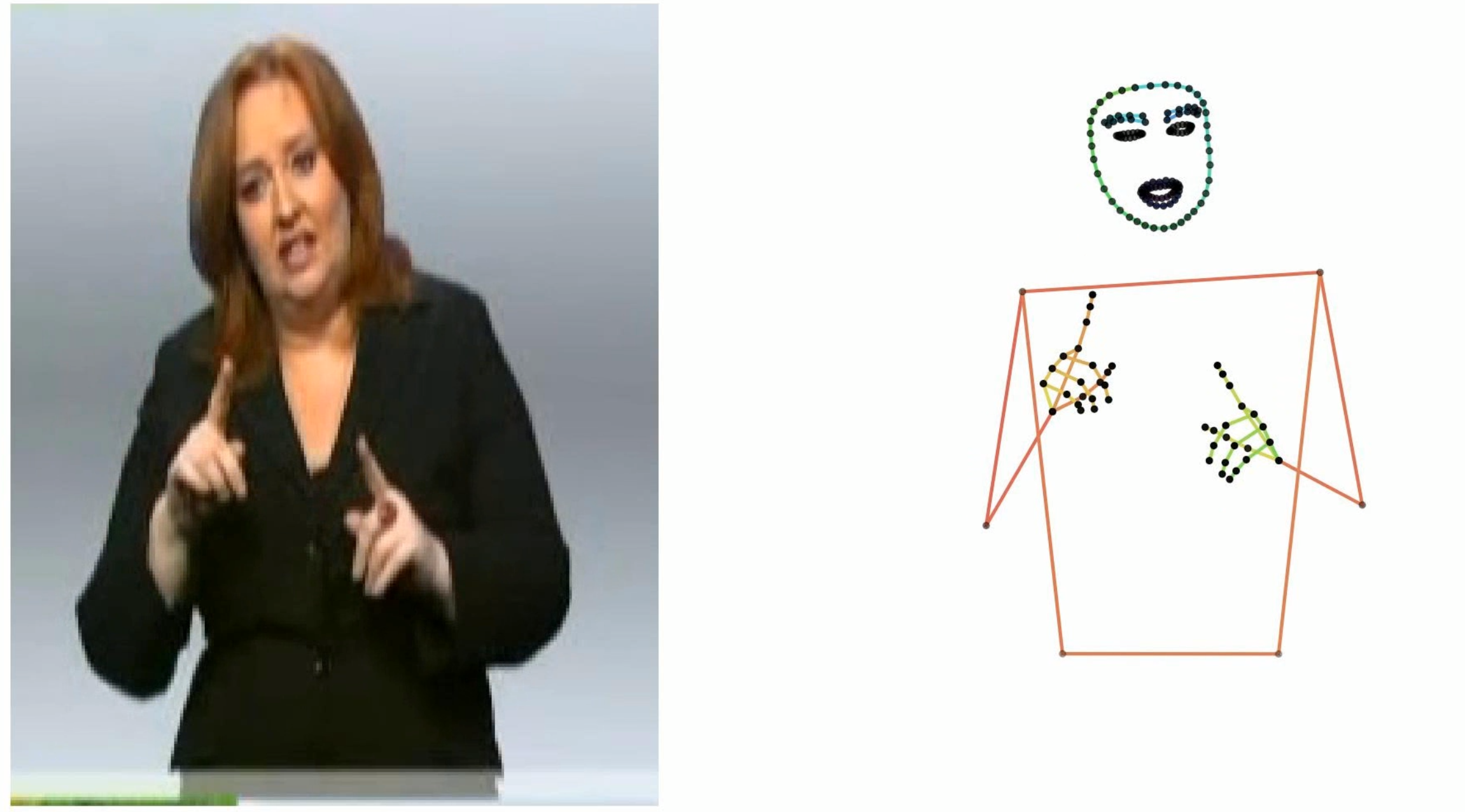}
    \caption{Skeleton extraction example, (left: original signer, right: 178 keypoint skeleton.}
    \label{fig:skeleton_example}
\end{figure}

For each video, we extract Mediapipe holistic keypoints and use the approach from Ivashechkin \etal
\cite{10193629}, to uplift the predictions to 3D. This optimization leverages a neural network, informed by human body physical constraints, to predict 3D skeleton joint angles from 2D keypoints. These angles are then used to apply a canonical skeleton, thereby ensuring consistent bone lengths across all signers. This provides 178 keypoint representation: 21 keypoints for each hand, 128 keypoints for the face, and, 8 keypoints for the body. The face is a subset of the 468 keypoint representations from Mediapipe, we do this for computational efficiency. We then normalise the skeleton such that the neck is at the origin and the body as is fixed on the xy plane. An example of the skeleton extraction can be seen in \cref{fig:skeleton_example}.

\section{Evaluation Metrics}
\label{sec:protocol}
Here we discuss the evaluation protocol used for the \ac{slp} challenge, which is also publicly available\footnote{Evaluation code: \url{https://github.com/walsharry/SLRTP-Sign-Production-Evaluation}}. The evaluation is comprised of text-based and pose-based metrics. Text-based metrics employ a back-translation model to convert the skeletal representation back into spoken language. Specifically, a ``Sign Language Transformer" \cite{camgoz2020sign} is utilized. The input layer is adapted to match the dimensionality of the keypoints. Both the encoder and the decoder component comprise three layers, 8 attention heads, with an embedding and feedforward dimensions of 256 and 512, respectively. Whereas the pose-based metric is calculated using the skeleton itself.

\subsection{Text-based}

\textbf{BLEU}: The Bilingual Evaluation Understudy (BLEU) score \cite{papineni2002bleu} compares a given sentence to that of a set of reference sentences by calculating the precision of word-level n-grams. Here we use n-grams one to four. 

\noindent
\textbf{CHRF}: The CHaRacter-level F-score (CHRF) \cite{popovic-2015-chrf} calculates the F-score based on the precision and recall of character-level n-grams between two sentences.

\noindent
\textbf{ROUGE}: The Recall-Oriented Understudy for Gisting Evaluation (ROUGE) score \cite{lin2004rouge} measures the overlap of n-grams, word sequences, and word pairs between two sentences.

\noindent
\textbf{WER}: The Word Error Rate (WER) calculates the number of errors (insertions, deletions, and substitutions) divided by the total number of words in the reference text.

\subsection{Pose-based}

\textbf{DTW MJE}: The Dynamic Time Warping Mean Joint Error (DTW MJE) is a metric used to evaluate the similarity between two sequences of skeleton poses. It calculates the average error between corresponding joints in the predicted and ground truth poses, considering temporal alignment.

\textbf{Total Distance}: This metric measures the overall distance the signer's hands have moved in 3D space. This is to judge how expressive the productions are. The prediction is normalized by the ground truth distance, therefore, a score of 1 is optimal.

\subsection{Baseline Model}

For a baseline method, we use the publicly available ``Progressive transformer" by Saunders \etal \cite{saunders2020progressive}. The model can translate from text to continuous 3D sign pose sequences in an end-to-end manner. The model is based on the transformer architecture, which has been shown to be effective for sequence-to-sequence tasks such as machine translation.  By introducing a novel counter decoding technique, the model can generate continuous pose sequences of variable lengths. 

We train a model for 300 epochs with an Adam optimizer with a learning rate of 0.001. In line with the original paper, both the encoder and decoder contain 2 layers and 8 heads with an embedding size of 512.
\section{Methodology}
\label{sec:methodology}

In this section, we introduce the three top-performing approaches in the SLRTP 2025 Sign Language Production Challenge. \emph{Team 1} employs a retrieval-based pipeline centered on gloss annotations, \emph{Team 2} leverages a generative diffusion-based model, and \emph{Team 3} presents a gloss-free transformer architecture with an autoencoder for latent pose embeddings. \cref{tab:general_info} summarizes the key characteristics of each method.

\begin{table}[htbp!]
\centering
\resizebox{0.49\textwidth}{!}{
\begin{tabular}{|l|c|c|c|} 
\hline
Participant                      & \begin{tabular}[c]{@{}c@{}}Team-1:\\USTC-MoE\end{tabular} & \begin{tabular}[c]{@{}c@{}}Team-2:\\hfut-lmc\end{tabular} & \begin{tabular}[c]{@{}c@{}}Team-3:\\Hacettepe\end{tabular}  \\ 
\hline
Model architecture               & \multicolumn{1}{c|}{Rule based}                           & \multicolumn{1}{c|}{Diffusion}                            & \multicolumn{1}{c|}{Transformer}                            \\ 
\hline
Pre-trained models               & \tmark                                                        & -                                                        & \tmark                                                          \\ 
\hline
No. Trainable Parameters         & 355.9M                                                    & 125.8M                                                         & 26M                                                         \\ 
\hline
External Datasets                & -                                                        & -                                                        & -                                                          \\ 
\hline
Gloss Information                & \tmark                                                        & -                                                        & -                                                          \\ 
\hline
Data Augmentation                & -                                                        & -                                                        & -                                                          \\ 
% \hline
% Skeleton Segmentation Techniques & -                                                        & -                                                        & -                                                          \\ 
\hline
Handcrafted Features             & -                                                        & -                                                        & -                                                          \\ 
\hline
Motion Constraints               & -                                                        & \tmark                                                        & -                                                          \\ 
% \hline
% Temporal Consistency             & -                                                        & -                                                        & -                                                          \\ 
\hline
Optimization Metric              & BLEU and DTW                                              & DTW                                                       & BLEU-4                                                      \\
\hline
\end{tabular}
}
\caption{General information about the top-3 winning approaches.}
\label{tab:general_info}
\end{table}

% \todo{maybe remove empty lines from general information table}

The following subsections detail each team's respective methodologies, including training setups and key design choices.

\begin{figure*}[t]
\centering
\centerline{\includegraphics[width=1\linewidth]{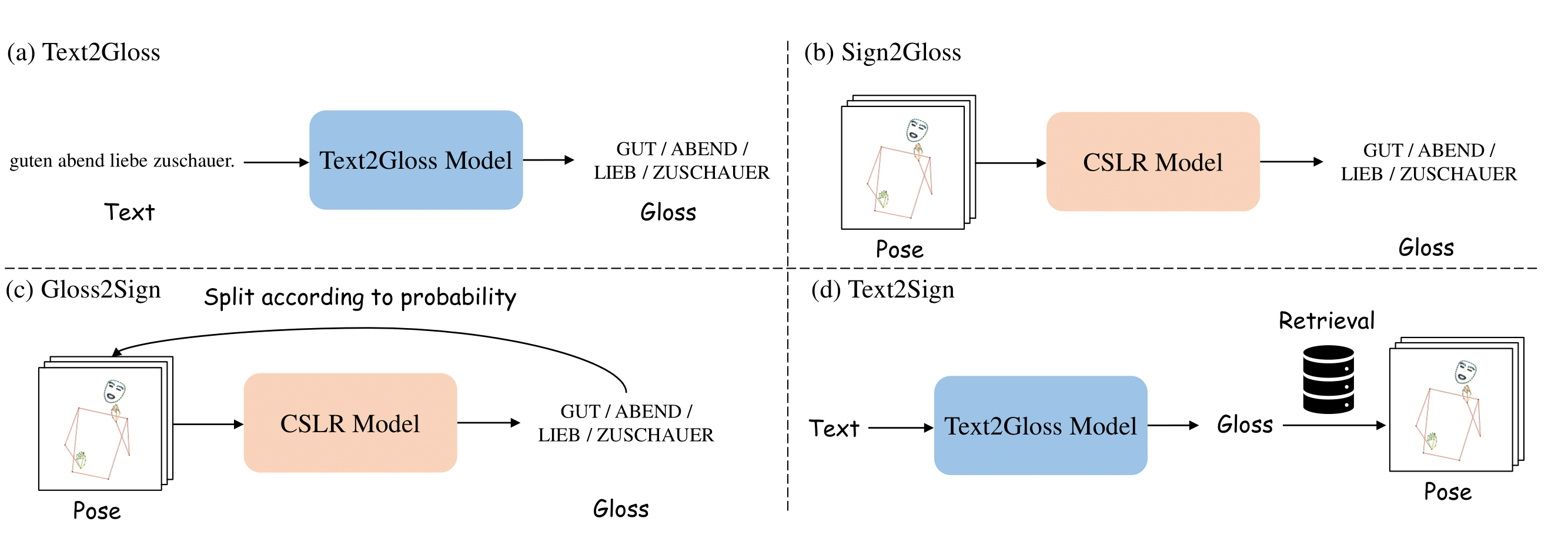}}
\caption{Overview figure of the 1\textsuperscript{st} place method (USTC-MoE).}
\label{fig:pipeline}
\end{figure*}

\subsection{Team 1: USTC-MoE (1\textsuperscript{st} Place)}
\noindent
\textbf{Overall Pipeline.}
\emph{Fig.~\ref{fig:pipeline}} (adapted from the team’s fact sheet) outlines the top-1 approach. It is a retrieval-based framework that connects spoken language input to a large dictionary of 3D pose segments. The method consists of four key modules: \emph{Text2Gloss}, \emph{Sign2Gloss}, \emph{Gloss-Pose Dictionary Construction}, and \emph{Sign Retrieval}.

\noindent
\textbf{(a) Text2Gloss.}
A multilingual pretrained language model, specifically XLM-R \cite{conneau2019unsupervised}, is employed to convert each spoken language sentence into a gloss sequence. This model is fine-tuned with a cosine annealing schedule for 40 epochs using Adam (weight decay $10^{-3}$, learning rate $10^{-5}$).

\noindent
\textbf{(b) Sign2Gloss.}
They then train a Continuous Sign Language Recognition (CSLR) model (based on TwoStream-SLT~\cite{chentwo}), which outputs a gloss label for every frame of the 3D pose sequence. This step provides a robust mapping from skeleton pose to discrete glosses.

\noindent
\textbf{(c) Gloss-Pose Dictionary Construction.}
After the CSLR network is trained, it is used to label and segment all training poses. Each gloss is therefore associated with its corresponding 3D sub-pose sequence, producing a large \emph{Gloss-Pose} dictionary of short signing segments.

\noindent
\textbf{(d) Text2Sign.}
At inference, the model (i)~translates text to gloss, (ii)~retrieves each sub-pose from the dictionary based on the gloss, and (iii)~concatenates these short pose segments to form the final sign pose sequence. 
They report a total parameter count of about 355.9\,M (dominated by the pretrained text model). No additional data augmentation is performed. The reliance on real pose segments ensures high fidelity and natural transitions across the retrieved signs.

\noindent
\textbf{Discussion.}
By grounding each gloss in an actual segment of human motion, this retrieval-based approach bypasses the complexities of motion synthesis and guarantees plausible, high-quality poses for individual signs. Concatenating these sub-pose sequences requires robust gloss alignment, achieved via the CSLR model. Its simplicity and reliability, coupled with strong text-to-gloss translation, led to first-place performance.

\vspace{2mm}
\subsection{Team 2: hfut-lmc (2\textsuperscript{nd} Place)}

\begin{figure}[!htb]
\centering
\includegraphics[width=1\columnwidth]{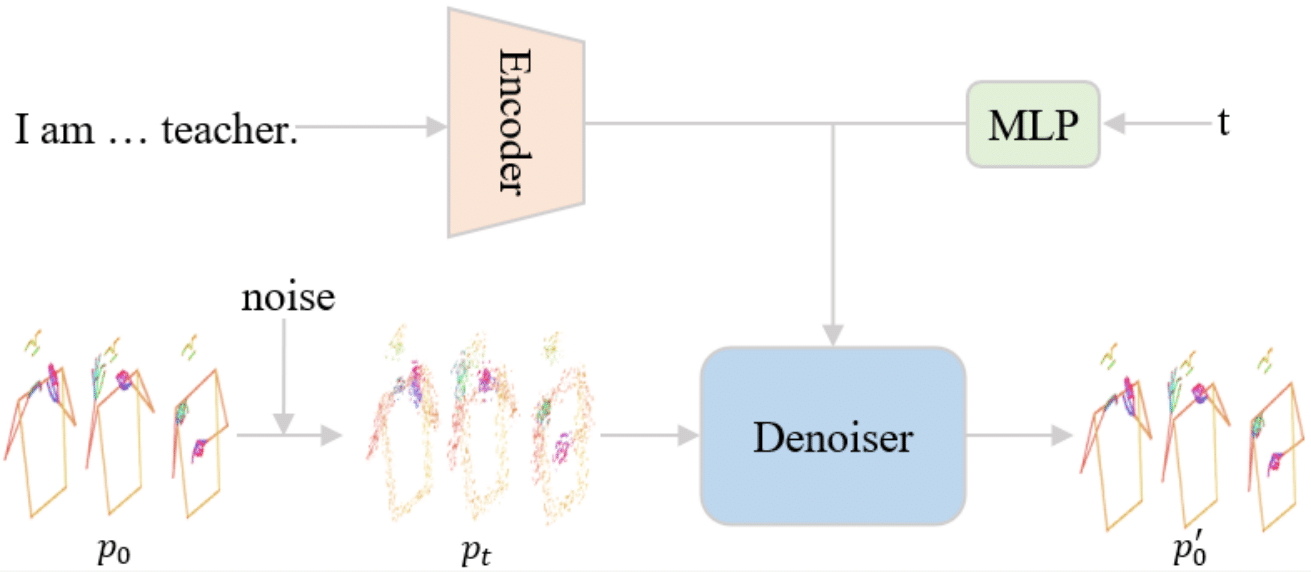}
\caption{Overall framework of the 2\textsuperscript{nd} place method (hfut-lmc).}
\label{fig: hfut_main}
\end{figure}

\noindent
\textbf{Overall Pipeline.}
Unlike the retrieval-based strategy from Team~1, hfut-lmc \cite{he2025text} proposes a fully generative diffusion-based framework, illustrated in \emph{Fig.~\ref{fig: hfut_main}}. 
Referred to as \emph{Text-Driven Conditional Diffusion Model (TCDM)}, their method learns an end-to-end mapping from textual input to sign language pose sequences without leveraging gloss-level supervision or large retrieval dictionaries.   

% A text encoder extracts high-level semantics, which are combined with a time-step embedding to guide the denoising process

\noindent
\textbf{Forward and Reverse Processes.}
Following the standard diffusion paradigm~\cite{Ho_Jain_Abbeel_Berkeley,Song_Meng_Ermon_2020}, the forward process gradually adds Gaussian noise to the ground-truth 3D pose sequence \(p_0\). 
After \(t\) steps, the noisy sequence \(p_t\) is:
\begin{equation}
    p_{t} \;=\;\gamma_{t}\,p_{0} \;+\;\sigma_{t}\,\epsilon,
    \label{eq: forward}
\end{equation}
where \(\epsilon\sim\mathcal{N}(0,1)\) and \(\gamma_{t}^2 + \sigma_{t}^2 = 1\). 
During training, the model learns a denoiser \(\mathcal{D}\bigl(p_t,\,g\bigr)\) that removes noise from \(p_t\), guided by a condition \(g\) derived from the text encoder plus the timestep \(t\):
\begin{equation}
    p_{0}' \;=\;\mathcal{D}\bigl(p_{t},\,g\bigr).
    \label{eq: reverse}
\end{equation}
At inference, an initially random pose \(p_{T}\) is iteratively denoised over a small number (5) of DDIM sampling steps, culminating in a coherent sign language motion.

\begin{figure}[t]
\centering
\includegraphics[width=0.45\columnwidth]{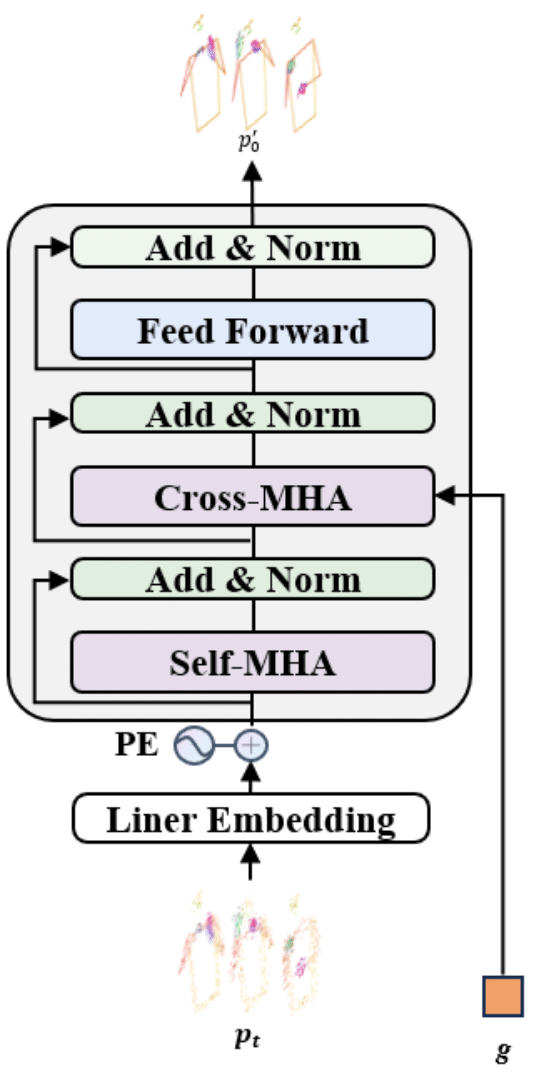}
\caption{Detailed implementation details of the denoiser $\mathcal{D}$, from the 2\textsuperscript{nd} place method (hfut-lmc).}
\label{fig: hfut_detail}
\end{figure}

\noindent
\textbf{Denoiser Architecture.}
To clarify how \(\mathcal{D}\) operates, hfut-lmc splits the procedure into several sub-stages (Fig.~\ref{fig: hfut_detail}):

\begin{enumerate}
    \item \emph{Linear Embedding Layer (LE):} The noisy pose \(p_{t}\) is projected into a higher-dimensional space:
    \begin{equation}
        p_{u} \;=\; W^{p}\,p_{t} \;+\; b^{p},
        \label{eq: linear}
    \end{equation}
    where \(p_{u}\) is the embedded representation of \(p_{t}\).
    
    \item \emph{Positional Encoding (PE):} A predefined sinusoidal encoding is added to inject temporal information:
    \begin{equation}
        \hat{p}_{u} \;=\; p_{u} \;+\; PE(n),
        \label{eq: posenc}
    \end{equation}
    where \(n\) indexes each frame in the sequence.

    \item \emph{Multi-Head Attention and Cross-Attention:} The embedded pose \(\hat{p}_{u}\) interacts with the conditioning \(g\) (derived from the text encoder and the current diffusion timestep) via cross-attention, yielding updated pose features that finally yield a refined pose estimate \(\hat{p}_{0}'\).
\end{enumerate}

\noindent
\textbf{Loss Functions.}
To ensure the generated poses are both accurate and realistic, hfut-lmc combines two complementary objectives:
\begin{itemize}
    \item \emph{Joint Position Loss} \(\mathcal{L}_{\text{joint}}\),  
    \begin{equation}
        \mathcal{L}_{\text{joint}} \;=\; \frac{1}{J}\sum_{j=1}^{J}\bigl\lVert p_{j} - p_{j}'\bigr\rVert,
        \label{eq: Ljoint}
    \end{equation}
    enforcing alignment between predicted and ground-truth joint coordinates.
    
    \item \emph{Bone Orientation Loss} \(\mathcal{L}_{\text{bone}}\), 
    \begin{equation}
       \mathcal{L}_{\text{bone}} \;=\; \frac{1}{B}\sum_{b=1}^{B}\bigl\lVert q_{b} - q_{b}'\bigr\rVert^{2},
       \label{eq: Lbone}
    \end{equation}
    promoting realistic limb orientations by comparing direction vectors for each bone.
\end{itemize}
The total training objective is
\begin{equation}
   \mathcal{L}_{\text{total}} \;=\; \mathcal{L}_{\text{joint}} \;+\; \lambda\,\mathcal{L}_{\text{bone}},
   \quad \lambda=0.1.
   \label{eq: Lfinal}
\end{equation}

\noindent
% \textbf{Implementation Details.}
They employ the text encoder architecture that is the same as the baseline method~\cite{saunders2020progressive}, but with 4 layers, 8 heads, and a 1024-dimensional embedding size. 
The forward diffusion steps is set to 1000 and is combined with a 5-step DDIM inference schedule. Training uses Adam~\cite{Kingma_Ba_2014} and a learning rate of $10^{-3}$.

\noindent
\textbf{Discussion.}
By omitting gloss supervision, hfut-lmc avoids the expense of additional annotations. Their bone-orientation and joint positioning losses, ensure the poses represent natural articulations, mitigating regression-to-mean artifacts.

\vspace{2mm}
\subsection{Team 3: Hacettepe (3\textsuperscript{rd} Place)}

\begin{figure}[!htb]
    \centering
    \includegraphics[width=0.9\columnwidth]{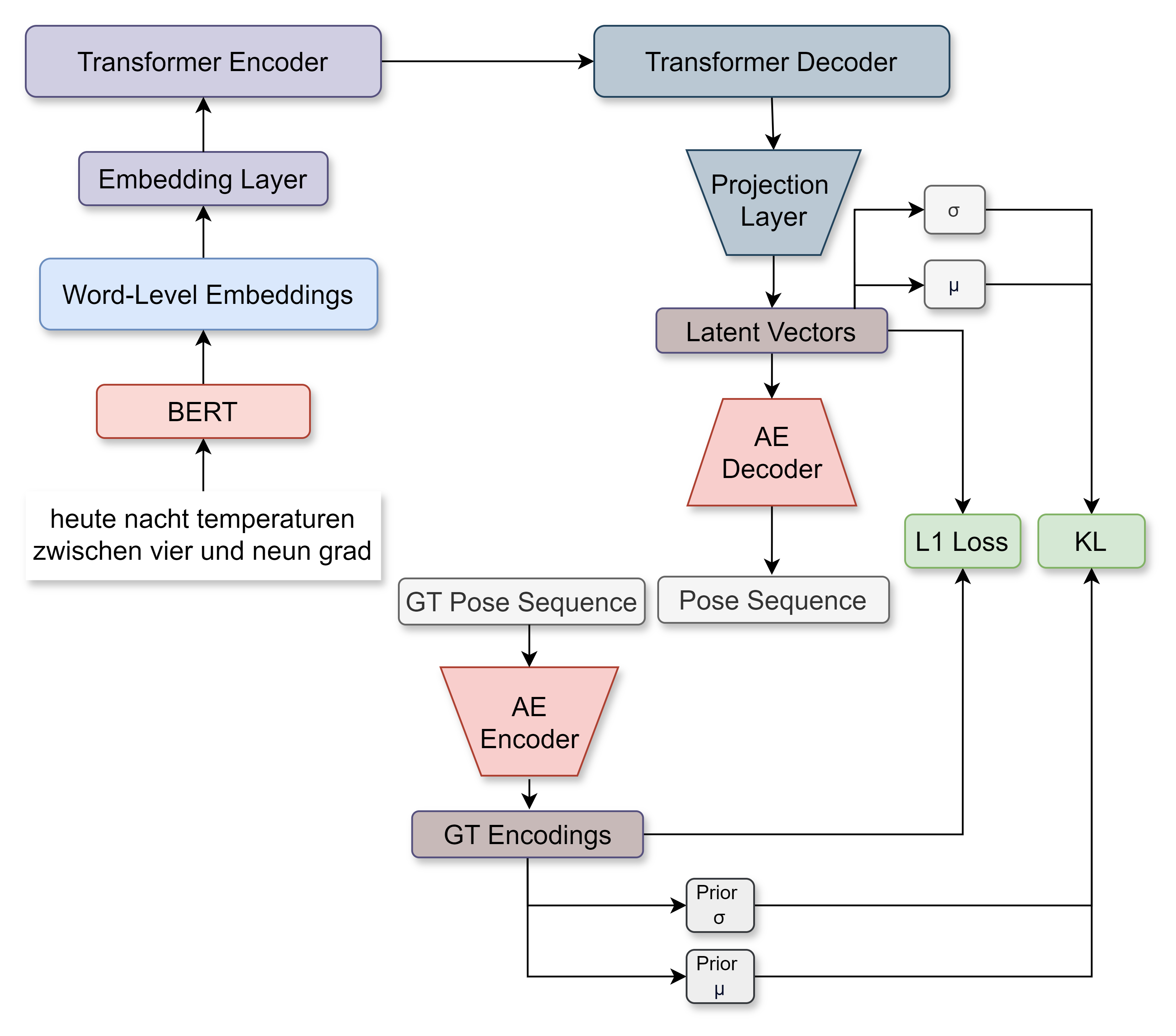}
    \caption{Workflow diagram for the 3\textsuperscript{rd} place method (Hacettepe).}
    \label{fig:hacettepe_main}
\end{figure}

\noindent
\textbf{Overall Pipeline.}
Hacettepe presents a \emph{gloss-free} transformer-based method that learns a compact, disentangled latent representation of sign pose sequences via an autoencoder. Channel-aware regularization guides text-to-pose mapping without gloss supervision, as shown in \emph{Fig.~\ref{fig:hacettepe_main}}.  

\begin{table*}[htbp!]
\centering
\resizebox{0.95\textwidth}{!}{
\begin{tabular}{r|ccccccccc} 
\toprule
\multicolumn{1}{l|}{}   & \multicolumn{9}{c}{RWTH-PHOENIX-Weather-2014\textbf{T} Test Set}                                                                                                            \\
Teams                   & BLEU-1         & BLEU-2         & BLEU-3        & BLEU-4        & CHRF           & ROUGE & WER $\downarrow$            & DTW-MJE $\downarrow$         & Total Distance  \\ 
\midrule
Ground Truth                & 34.40 & 22.04 & 16.09 & 12.78 & 34.59 & 35.20 & 85.77 & 0.0000 & 1.000 \\
\midrule
Team 1 (USTC-MoE)                     & \textbf{34.85} & \textbf{21.96} & \textbf{15.65} & \textbf{12.06} & \textbf{36.83} & \textbf{36.59} & 93.49 & 0.0448 & 1.631 \\
Team 2 (hfut-lmc)                      & 16.96 & 6.56 & 3.38 & 2.05 & 25.88 & 19.77 & 147.85 & \textbf{0.0403} & 2.512 \\ 
Team 3 (Hacettepe)                     & 30.44 & 17.75 & 12.42 & 9.59  & 29.70 & 30.64 & \textbf{88.88} & 0.0423 & \textbf{0.798} \\ 
\hline
Progressive Transformer \cite{saunders2020progressive}    & 22.17 & 10.71 & 7.09 & 5.43 & 24.13 & 21.98 & 101.45 & 0.0418 & 0.257 \\ 
\bottomrule
\end{tabular}
}
\caption{SLP Challenge Results on the \acf{ph14t} test set. }
\label{table:phix_test_metrics}
\end{table*}

\begin{table*}[htbp!]
\centering
\resizebox{0.95\textwidth}{!}{
\begin{tabular}{r|ccccccccc} 
\toprule
\multicolumn{1}{l|}{}   & \multicolumn{9}{c}{Hidden Test Set}                                                                                                            \\
Teams                   & BLEU-1         & BLEU-2         & BLEU-3        & BLEU-4        & CHRF           & ROUGE & WER  $\downarrow$           & DTW-MJE $\downarrow$         & Total Distance  \\ 
\midrule
Ground Truth            & 37.94 & 19.87 & 10.67 & 5.90 & 30.64 & 38.60 & 101.25 & 0.0000 & 1.000 \\ 
\midrule
Team 1 (USTC-MoE)       & \textbf{31.40} & \textbf{17.09} & \textbf{9.43} & \textbf{5.86} & \textbf{31.73} & \textbf{33.75} & 109.38 & 0.0574 & 1.185  \\
Team 2 (hfut-lmc)       & 30.54 & 16.22 & 9.33 & 5.66 & 30.17 & 32.92 & 107.93 & \textbf{0.0492} & \textbf{0.971}  \\ 
Team 3 (Hacettepe)      & 27.51 & 11.13 & 5.36 & 2.91 & 23.37 & 27.29 & \textbf{105.49} & 0.0531 & 0.761  \\
\hline
Progressive Transformer \cite{saunders2020progressive} & 18.33 & 4.99 & 1.74 & 0.78 & 21.65 & 21.10 & 141.93 & 0.0467 & 0.322           \\
\bottomrule
\end{tabular}
}
\caption{SLP Challenge Results on the Hidden Test Set. }
\label{table:secret_test_metrics}
\end{table*}

\noindent
\textbf{Autoencoder for Latent Pose.}
A structurally disentangled pose autoencoder
is first pre-trained to reconstruct 3D poses. Each 534-dimensional input is factorized into four articulatory regions, face, body, left hand, and right hand, mapped to an 80-dimensional
latent space. Region-specific L1 reconstruction and encoder-weight regularization ensure
compact and semantically structured representations.

\noindent
\textbf{Transformer Architecture.}
A non-autoregressive transformer then predicts these latent embeddings from 768-dimensional sentence vectors obtained from a pretrained BERT
model. Text embeddings are reduced to 512 dimensions before being processed by a 3-layer encoder and a 6-layer decoder to generate pose sequences in parallel, matching the autoencoder’s latent space.

\noindent
\textbf{Reconstruction \& Training.} The transformer is trained with an L1 loss between predicted latent vectors and the autoencoder’s ground-truth codes, ensuring realistic 3D motion reconstruction. In the second phase, channel-wise KL divergence promotes articulator-aware regularization. As it avoids gloss annotations, the model scales efficiently without costly linguistic labelling.

\noindent
\textbf{Discussion.} 
Hacettepe’s gloss-free, disentangled autoencoder-based approach offers a compact, interpretable representation for SLP. Unlike retrieval-based methods, it synthesizes entirely new sequences without gloss or a motion dictionary. Although bridging text to a learned pose space can present challenges for highly nuanced signs, the method demonstrates competitive accuracy and efficiency, securing third place in the challenge. Further details and extended results are provided in
\cite{tasyurek2025disentangle}.
\section{Challenge Results}
\label{sec:results}

On the hidden test set, all three teams significantly outperformed the baseline method.  The top-performing team achieved an increase of 13.07 in BLEU-1, as shown in \cref{table:secret_test_metrics}. A key limitation noted by the baseline method was the issue of regression to the mean. This observation is quantified by the proposed total distance metric, which reveals that the progressive transformer generated motion that traversed only 25\% of the ground truth distance. In contrast, all three teams produced more expressive outputs that more closely approximated the ground truth motion.

Team 2 outperformed the baseline on the hidden test set. However, on the \ac{ph14t} test set, its performance decreased substantially, falling below the baseline. To understand this performance disparity between the two test sets, we analyzed the predicted sequences. We found that the average length of Team 2's predictions was 2.463 times longer than the ground truth on the \ac{ph14t} test set, whereas on the hidden test set, it was close to 1. In comparison, Teams 1, 3, and the baseline exhibited average duration ratios of 1.438, 1.026, and 0.999, respectively. Just as spoken languages employ intonation, rhythm, and stress to convey nuance and meaning, sign languages utilize prosody. Thus, losing this information and producing longer sequences can be damaging to the performance. We hypothesize that Team 3's and the baseline's transformer-based architecture, ideally suited for sequence-to-sequence tasks, effectively captured this prosodic information. 

Discrepancies emerged between certain pose-based and text-based metrics. Specifically, DTW-MJE tended to favor methods that produced less articulated or longer sequences. This observation underscores the importance of employing a diverse range of evaluation metrics when assessing generative models.

\begin{figure*}[htbp!]
\centering
\includegraphics[width=0.98\textwidth, trim=165 50 165 0, clip]{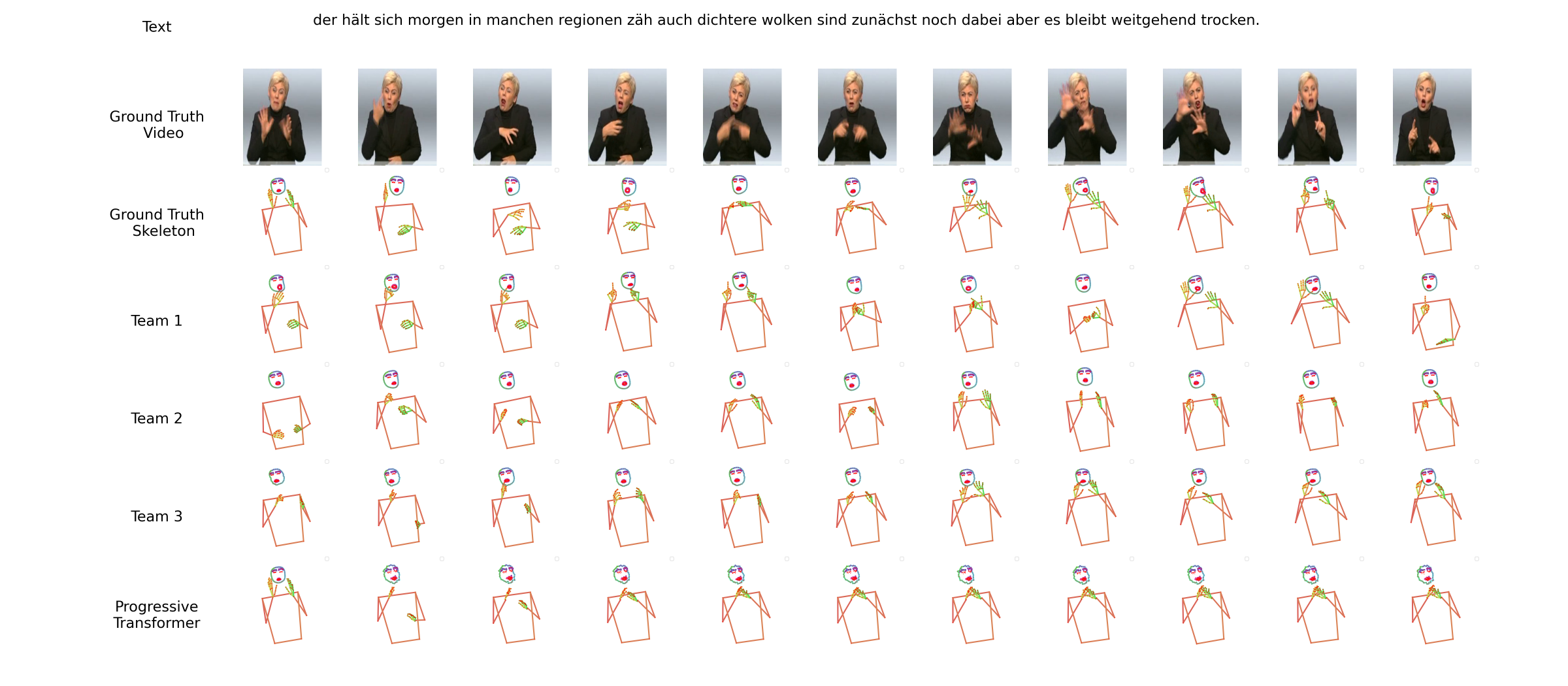}
\caption{Qualatative results of the top 3 performing teams on the \ac{ph14t} test set. The first row shows the input spoken language sentence, the second and third rows show the ground truth video and skeleton, respectively, and the subsequent rows show the predicted skeleton poses from the top 3 teams. The final row shows the progressive transformer baseline.}
\label{results:qualitative}
\end{figure*}

Across both test sets, Team 3 consistently produced the lowest Word Error Rate (WER). Notably, this approach incorporated BERT, a model pre-trained specifically on German. Team 1, which fine-tuned XLM-R, a multilingual model, demonstrated superior translation quality, as evidenced by higher BLEU scores. Given the low-resource nature of sign language, we hypothesize that the model's pre-existing linguistic knowledge significantly contributed to this performance. This suggests that spoken language resources can be effectively leveraged for \ac{slp}.

A correlation was observed between team rankings and model size; Team 1's model contained 13.7 times more parameters than Team 3's, and 2.8 times more than Team 2's.

Decomposing the WER metric into its constituent components revealed that each team predominantly exhibited errors within a distinct category. Specifically, Team 1 encountered the most replacement errors, Team 2 the most deletion errors, and Team 3 the most insertion errors. The hidden test set exhibited an average sentence length of eight words, with a range from three to fifteen words. A negative correlation was observed, on average, between sentence length and error rate across all three teams. This suggests a potential tendency towards over-translation, although this effect could originate from either the back-translation model or the teams' respective approaches, a distinction that is challenging to isolate. Overall, the words `und,' `regen,' and `süden' were identified as the most frequent sources of error in the translations.

\textbf{Qualitative results:} Evaluation of the qualitative results supports the previous findings. The impact of the increased BLEU-1 score, noted for Team 1, is demonstrated in \cref{results:qualitative}. This retrieval-based approach ensures that each generated sequence exhibits expressive signing. This observation is further quantified by our proposed total distance metric. As presented in Table \ref{table:secret_test_metrics}, Team 1 was the only team to achieve a score surpassing one. Overall, the outputs from all teams preserved features of the ground truth sequences. However, further research is necessary to achieve the fluency and naturalness of native Deaf signers.

\section{Conclusion}
\label{sec:conclusion}
This paper summarizes the findings of the first \ac{slp} challenge. In the final test phase, six teams submitted solutions that outperformed the baseline approach. We presented the work of the top three teams. The top-performing team utilized a recognition model to curate a dictionary of signs, which was subsequently employed for production. We note that gloss annotations are a limited resource and therefore, would limit scaling this method to larger datasets. Teams 2 and 3 circumvented the requirement for this type of annotation and presented models that demonstrated competitive performance despite the absence of linguistic annotation.

Evaluating sign language is a challenging task that is compounded by the fact that data normalisations and representations are constantly evolving. We hope this work helps to establish a more consistent baseline for future SLP research. We implore future researchers to release their process features for other data sets, as without consistent inputs, comparison between approaches is meaningless. 
\section*{Acknowledgments}
\label{sec:Acknowledgemnts}
This work was supported by the SNSF project ‘SMILE II’ (CRSII5 193686), the Innosuisse IICT Flagship (PFFS-21-47), EPSRC grant APP24554 (SignGPT-EP/Z535370/1), Google DeepMind and through funding from Google.org via the AI for Global Goals scheme. This work reflects only the author’s views and the funders are not responsible for any use that may be made of the information it contains.

{
    \small
    \bibliographystyle{ieeenat_fullname}
    \bibliography{main}

\begin{thebibliography}{55}
\providecommand{\natexlab}[1]{#1}
\providecommand{\url}[1]{\texttt{#1}}
\expandafter\ifx\csname urlstyle\endcsname\relax
  \providecommand{\doi}[1]{doi: #1}\else
  \providecommand{\doi}{doi: \begingroup \urlstyle{rm}\Url}\fi

\bibitem[Baltatzis et~al.(2024)Baltatzis, Potamias, Ververas, Sun, Deng, and Zafeiriou]{baltatzis2024neural}
Vasileios Baltatzis, Rolandos~Alexandros Potamias, Evangelos Ververas, Guanxiong Sun, Jiankang Deng, and Stefanos Zafeiriou.
\newblock Neural sign actors: a diffusion model for 3d sign language production from text.
\newblock In \emph{Proceedings of the IEEE/CVF Conference on Computer Vision and Pattern Recognition}, pages 1985--1995, 2024.

\bibitem[Bangham et~al.(2000)Bangham, Cox, Elliott, Glauert, Marshall, Rankov, and Wells]{bangham2000virtual}
Andrew Bangham, SJ Cox, Ralph Elliott, John~RW Glauert, Ian Marshall, Sanja Rankov, and Mark Wells.
\newblock Virtual signing: Capture, animation, storage and transmission-an overview of the visicast project.
\newblock In \emph{IEE Seminar on speech and language processing for disabled and elderly people (Ref. No. 2000/025)}, pages 6--1. IET, 2000.

\bibitem[Bungeroth and Ney(2004)]{Bungeroth2004StatisticalSL}
Jan Bungeroth and Hermann Ney.
\newblock Statistical sign language translation.
\newblock In \emph{sign-lang@ LREC 2004}, pages 105--108. Citeseer, 2004.

\bibitem[Camgoz et~al.(2018)Camgoz, Hadfield, Koller, Ney, and Bowden]{camgoz2018neural}
Necati~Cihan Camgoz, Simon Hadfield, Oscar Koller, Hermann Ney, and Richard Bowden.
\newblock Neural sign language translation.
\newblock In \emph{Proceedings of the IEEE conference on computer vision and pattern recognition}, pages 7784--7793, 2018.

\bibitem[Camgoz et~al.(2020)Camgoz, Koller, Hadfield, and Bowden]{camgoz2020sign}
Necati~Cihan Camgoz, Oscar Koller, Simon Hadfield, and Richard Bowden.
\newblock Sign language transformers: Joint end-to-end sign language recognition and translation.
\newblock In \emph{Proceedings of the IEEE/CVF conference on computer vision and pattern recognition}, pages 10023--10033, 2020.

\bibitem[Cao et~al.(2019)Cao, Hidalgo, Simon, Wei, and Sheikh]{cao2019openposerealtimemultiperson2d}
Zhe Cao, Gines Hidalgo, Tomas Simon, Shih-En Wei, and Yaser Sheikh.
\newblock Openpose: Realtime multi-person 2d pose estimation using part affinity fields, 2019.

\bibitem[Chen et~al.(2024)Chen, Wang, and Wang]{CHEN2024104050}
Sheng Chen, Qingshan Wang, and Qi Wang.
\newblock Semantic-driven diffusion for sign language production with gloss-pose latent spaces alignment.
\newblock \emph{Computer Vision and Image Understanding}, 246:\penalty0 104050, 2024.

\bibitem[Chen et~al.(2022)Chen, Zuo, Wei, Wu, Liu, and Mak]{chentwo}
Yutong Chen, Ronglai Zuo, Fangyun Wei, Yu Wu, Shujie Liu, and Brian Mak.
\newblock Two-stream network for sign language recognition and translation.
\newblock In \emph{NeurIPS}, 2022.

\bibitem[Conneau et~al.(2019)Conneau, Khandelwal, Goyal, Chaudhary, Wenzek, Guzm{\'a}n, Grave, Ott, Zettlemoyer, and Stoyanov]{conneau2019unsupervised}
Alexis Conneau, Kartikay Khandelwal, Naman Goyal, Vishrav Chaudhary, Guillaume Wenzek, Francisco Guzm{\'a}n, Edouard Grave, Myle Ott, Luke Zettlemoyer, and Veselin Stoyanov.
\newblock Unsupervised cross-lingual representation learning at scale.
\newblock \emph{arXiv preprint arXiv:1911.02116}, 2019.

\bibitem[Cox et~al.(2002)Cox, Lincoln, Tryggvason, Nakisa, Wells, Tutt, and Abbott]{cox2002tessa}
Stephen Cox, Michael Lincoln, Judy Tryggvason, Melanie Nakisa, Mark Wells, Marcus Tutt, and Sanja Abbott.
\newblock Tessa, a system to aid communication with deaf people.
\newblock In \emph{Proceedings of the fifth international ACM conference on Assistive technologies}, pages 205--212, 2002.

\bibitem[Efthimiou et~al.(2012)Efthimiou, Fotinea, Hanke, Glauert, Bowden, Braffort, Collet, Maragos, and Lefebvre-Albaret]{efthimiou2012dicta}
Eleni Efthimiou, Stavroula-Evita Fotinea, Thomas Hanke, John Glauert, Richard Bowden, Annelies Braffort, Christophe Collet, Petros Maragos, and Fran{\c{c}}ois Lefebvre-Albaret.
\newblock The dicta-sign wiki: Enabling web communication for the deaf.
\newblock In \emph{International Conference on Computers for Handicapped Persons}, pages 205--212. Springer, 2012.

\bibitem[ElGhoul and Jemni(2011)]{elghoul2011websign}
Oussama ElGhoul and Mohamed Jemni.
\newblock Websign: A system to make and interpret signs using 3d avatars.
\newblock In \emph{Proceedings of the Second International Workshop on Sign Language Translation and Avatar Technology (SLTAT), Dundee, UK}, 2011.

\bibitem[Fang et~al.(2023)Fang, Sui, Zhang, and Tian]{fang2023signdifflearningdiffusionmodels}
Sen Fang, Chunyu Sui, Xuedong Zhang, and Yapeng Tian.
\newblock Signdiff: Learning diffusion models for american sign language production, 2023.

\bibitem[Gibet et~al.(2016)Gibet, Lefebvre-Albaret, Hamon, Brun, and Turki]{gibet2016interactive}
Sylvie Gibet, Fran{\c{c}}ois Lefebvre-Albaret, Ludovic Hamon, R{\'e}mi Brun, and Ahmed Turki.
\newblock Interactive editing in french sign language dedicated to virtual signers: Requirements and challenges.
\newblock \emph{Universal Access in the Information Society}, 15:\penalty0 525--539, 2016.

\bibitem[Glauert et~al.(2006)Glauert, Elliott, Cox, Tryggvason, and Sheard]{glauert2006vanessa}
John~RW Glauert, Ralph Elliott, Stephen~J Cox, Judy Tryggvason, and Mary Sheard.
\newblock Vanessa--a system for communication between deaf and hearing people.
\newblock \emph{Technology and disability}, 18\penalty0 (4):\penalty0 207--216, 2006.

\bibitem[Guo et~al.(2024)Guo, He, Jiao, Wang, Wang, Chen, Tu, Xu, and Zhang]{guo2024unsupervisedsignlanguagetranslation}
Zhengsheng Guo, Zhiwei He, Wenxiang Jiao, Xing Wang, Rui Wang, Kehai Chen, Zhaopeng Tu, Yong Xu, and Min Zhang.
\newblock Unsupervised sign language translation and generation, 2024.

\bibitem[He et~al.(2025)He, Wang, Zhang, Tang, Wang, and Cheng]{he2025text}
Jiayi He, Xu Wang, Ruobei Zhang, Shengeng Tang, Yaxiong Wang, and Lechao Cheng.
\newblock Text-driven diffusion model for sign language production.
\newblock \emph{arXiv preprint arXiv:2503.15914}, 2025.

\bibitem[Ho et~al.(2020)Ho, Jain, and Abbeel]{Ho_Jain_Abbeel_Berkeley}
Jonathan Ho, Ajay Jain, and Pieter Abbeel.
\newblock Denoising diffusion probabilistic models.
\newblock \emph{Advances in Neural Information Processing Systems}, pages 6840--6851, 2020.

\bibitem[Huang et~al.(2021)Huang, Pan, Zhao, and Tian]{huang2021towards}
Wencan Huang, Wenwen Pan, Zhou Zhao, and Qi Tian.
\newblock Towards fast and high-quality sign language production.
\newblock In \emph{Proceedings of the 29th ACM International Conference on Multimedia}, pages 3172--3181, 2021.

\bibitem[Hwang et~al.(2023)Hwang, Lee, and Park]{hwang2023autoregressive}
Eui~Jun Hwang, Huije Lee, and Jong~C Park.
\newblock Autoregressive sign language production: A gloss-free approach with discrete representations.
\newblock \emph{arXiv preprint arXiv:2309.12179}, 2023.

\bibitem[Ivashechkin et~al.(2023)Ivashechkin, Mendez, and Bowden]{10193629}
Maksym Ivashechkin, Oscar Mendez, and Richard Bowden.
\newblock Improving 3d pose estimation for sign language.
\newblock In \emph{2023 IEEE International Conference on Acoustics, Speech, and Signal Processing Workshops (ICASSPW)}, pages 1--5, 2023.

\bibitem[Kanis et~al.(2006)Kanis, Zahradil, Jur{\v{c}}{\'\i}{\v{c}}ek, and M{\"u}ller]{kanis2006czech}
Jakub Kanis, Ji{\v{r}}{\'\i} Zahradil, Filip Jur{\v{c}}{\'\i}{\v{c}}ek, and Lud{\v{e}}k M{\"u}ller.
\newblock Czech-sign speech corpus for semantic based machine translation.
\newblock In \emph{International Conference on Text, Speech and Dialogue}, pages 613--620. Springer, 2006.

\bibitem[Kingma and Ba(2015)]{Kingma_Ba_2014}
Diederik~P Kingma and Jimmy Ba.
\newblock Adam: A method for stochastic optimization.
\newblock In \emph{International Conference on Learning Representations}, pages 1--15, 2015.

\bibitem[Lin(2004)]{lin2004rouge}
Chin-Yew Lin.
\newblock Rouge: A package for automatic evaluation of summaries.
\newblock In \emph{Text summarization branches out}, pages 74--81, 2004.

\bibitem[Loper et~al.(2015)Loper, Mahmood, Romero, Pons-Moll, and Black]{loper2015smpl}
Matthew Loper, Naureen Mahmood, Javier Romero, Gerard Pons-Moll, and Michael~J Black.
\newblock Smpl: A skinned multi-person linear model.
\newblock \emph{TOG}, 34\penalty0 (6):\penalty0 1--16, 2015.

\bibitem[Lugaresi et~al.(2019)Lugaresi, Tang, Nash, McClanahan, Uboweja, Hays, Zhang, Chang, Yong, Lee, et~al.]{lugaresi2019mediapipe}
Camillo Lugaresi, Jiuqiang Tang, Hadon Nash, Chris McClanahan, Esha Uboweja, Michael Hays, Fan Zhang, Chuo-Ling Chang, Ming~Guang Yong, Juhyun Lee, et~al.
\newblock Mediapipe: A framework for building perception pipelines.
\newblock \emph{arXiv preprint arXiv:1906.08172}, 2019.

\bibitem[Othman and Jemni(2011)]{othman2011statisticalsignlanguagemachine}
Achraf Othman and Mohamed Jemni.
\newblock Statistical sign language machine translation: from english written text to american sign language gloss, 2011.

\bibitem[Papineni et~al.(2002)Papineni, Roukos, Ward, and Zhu]{papineni2002bleu}
Kishore Papineni, Salim Roukos, Todd Ward, and Wei-Jing Zhu.
\newblock Bleu: a method for automatic evaluation of machine translation.
\newblock In \emph{Proceedings of the 40th annual meeting of the Association for Computational Linguistics}, pages 311--318, 2002.

\bibitem[Pelykh et~al.(2024)Pelykh, Sincan, and Bowden]{pelykh2024givinghanddiffusionmodels}
Anton Pelykh, Ozge~Mercanoglu Sincan, and Richard Bowden.
\newblock Giving a hand to diffusion models: a two-stage approach to improving conditional human image generation, 2024.

\bibitem[Popovi{\'c}(2015)]{popovic-2015-chrf}
Maja Popovi{\'c}.
\newblock chr{F}: character n-gram {F}-score for automatic {MT} evaluation.
\newblock In \emph{Proceedings of the Tenth Workshop on Statistical Machine Translation}, pages 392--395, Lisbon, Portugal, 2015. Association for Computational Linguistics.

\bibitem[Saunders et~al.(2020{\natexlab{a}})Saunders, Camg{\"o}z, and Bowden]{saunders2020adversarial}
Ben Saunders, Necati~Cihan Camg{\"o}z, and Richard Bowden.
\newblock Adversarial training for multi-channel sign language production.
\newblock In \emph{British Machine Vision Virtual Conference}, 2020{\natexlab{a}}.

\bibitem[Saunders et~al.(2020{\natexlab{b}})Saunders, Camgoz, and Bowden]{saunders2020everybody}
Ben Saunders, Necati~Cihan Camgoz, and Richard Bowden.
\newblock Everybody sign now: Translating spoken language to photo realistic sign language video.
\newblock \emph{arXiv preprint arXiv:2011.09846}, 2020{\natexlab{b}}.

\bibitem[Saunders et~al.(2020{\natexlab{c}})Saunders, Camgoz, and Bowden]{saunders2020progressive}
Ben Saunders, Necati~Cihan Camgoz, and Richard Bowden.
\newblock Progressive transformers for end-to-end sign language production.
\newblock In \emph{Computer Vision--ECCV 2020: 16th European Conference, Glasgow, UK, August 23--28, 2020, Proceedings, Part XI 16}, pages 687--705. Springer, 2020{\natexlab{c}}.

\bibitem[Saunders et~al.(2021)Saunders, Camgoz, and Bowden]{saunders2021mixed}
Ben Saunders, Necati~Cihan Camgoz, and Richard Bowden.
\newblock Mixed signals: Sign language production via a mixture of motion primitives.
\newblock In \emph{Proceedings of the IEEE/CVF International Conference on Computer Vision}, 2021.

\bibitem[Saunders et~al.(2022{\natexlab{a}})Saunders, Camgoz, and Bowden]{saunders2021signing}
Ben Saunders, Necati~Cihan Camgoz, and Richard Bowden.
\newblock Signing at scale: Learning to co-articulate signs for large-scale photo-realistic sign language production.
\newblock In \emph{Proceedings of the IEEE Conference on Computer Vision and Pattern Recognition (CVPR)}, 2022{\natexlab{a}}.

\bibitem[Saunders et~al.(2022{\natexlab{b}})Saunders, Camgoz, and Bowden]{saunders2022signing}
Ben Saunders, Necati~Cihan Camgoz, and Richard Bowden.
\newblock Signing at scale: Learning to co-articulate signs for large-scale photo-realistic sign language production.
\newblock In \emph{Proceedings of the IEEE/CVF Conference on Computer Vision and Pattern Recognition}, pages 5141--5151, 2022{\natexlab{b}}.

\bibitem[Song et~al.(2020)Song, Meng, and Ermon]{Song_Meng_Ermon_2020}
Jiaming Song, Chenlin Meng, and Stefano Ermon.
\newblock Denoising diffusion implicit models.
\newblock In \emph{International Conference on Learning Representations}, 2020.

\bibitem[Stoll et~al.(2018)Stoll, Camg{\"o}z, Hadfield, and Bowden]{stoll2018sign}
Stephanie Stoll, Necati~Cihan Camg{\"o}z, Simon Hadfield, and R. Bowden.
\newblock Sign language production using neural machine translation and generative adversarial networks.
\newblock In \emph{British Machine Vision Conference}, 2018.

\bibitem[Stoll et~al.(2020)Stoll, Camgoz, Hadfield, and Bowden]{stoll2020text2sign}
Stephanie Stoll, Necati~Cihan Camgoz, Simon Hadfield, and Richard Bowden.
\newblock Text2sign: towards sign language production using neural machine translation and generative adversarial networks.
\newblock \emph{IJCV}, 128\penalty0 (4):\penalty0 891--908, 2020.

\bibitem[Stoll et~al.(2022)Stoll, Mustafa, and Guillemaut]{stoll2022there}
Stephanie Stoll, Armin Mustafa, and Jean-Yves Guillemaut.
\newblock There and back again: 3d sign language generation from text using back-translation.
\newblock In \emph{2022 International Conference on 3D Vision (3DV)}, pages 187--196. IEEE, 2022.

\bibitem[Sutton-Spence and Woll(1999)]{sutton1999linguistics}
Rachel Sutton-Spence and Bencie Woll.
\newblock \emph{The linguistics of British Sign Language: an introduction}.
\newblock Cambridge University Press, 1999.

\bibitem[Tamura and Kawasaki(1988)]{tamura1988recognition}
Shinichi Tamura and Shingo Kawasaki.
\newblock Recognition of sign language motion images.
\newblock \emph{Pattern Recognition}, 1988.

\bibitem[Tang et~al.(2024{\natexlab{a}})Tang, He, Cheng, Wu, Guo, and Hong]{tang2024discrete}
Shengeng Tang, Jiayi He, Lechao Cheng, Jingjing Wu, Dan Guo, and Richang Hong.
\newblock Discrete to continuous: Generating smooth transition poses from sign language observation.
\newblock \emph{arXiv preprint arXiv:2411.16810}, 2024{\natexlab{a}}.

\bibitem[Tang et~al.(2024{\natexlab{b}})Tang, Xue, Wu, Wang, and Hong]{10.1145/3663572}
Shengeng Tang, Feng Xue, Jingjing Wu, Shuo Wang, and Richang Hong.
\newblock Gloss-driven conditional diffusion models for sign language production.
\newblock \emph{ACM Trans. Multimedia Comput. Commun. Appl.}, 2024{\natexlab{b}}.
\newblock Just Accepted.

\bibitem[Taşyürek et~al.(2025)Taşyürek, Kızıltepe, and Keles]{tasyurek2025disentangle}
Sümeyye~Meryem Taşyürek, Tuğçe Kızıltepe, and Hacer~Yalim Keles.
\newblock Disentangle and regularize: Sign language production with articulator-based disentanglement and channel-aware regularization.
\newblock \emph{arXiv preprint arXiv:2504.06610}, 2025.

\bibitem[Walsh et~al.(2022)Walsh, Saunders, and Bowden]{walsh2022changing}
Harry Walsh, Ben Saunders, and Richard Bowden.
\newblock Changing the representation: Examining language representation for neural sign language production.
\newblock In \emph{LREC 2022 Workshop Language Resources and Evaluation Conference 24 June 2022}, page 117, 2022.

\bibitem[Walsh et~al.(2024{\natexlab{a}})Walsh, Ravanshad, Rahmani, and Bowden]{walsh2024data}
Harry Walsh, Abolfazl Ravanshad, Mariam Rahmani, and Richard Bowden.
\newblock A data-driven representation for sign language production.
\newblock In \emph{Proceedings of the 18th International Conference on Automatic Face and Gesture Recognition (FG 2024)}. Institute of Electrical and Electronics Engineers (IEEE), 2024{\natexlab{a}}.

\bibitem[Walsh et~al.(2024{\natexlab{b}})Walsh, Saunders, and Bowden]{walsh2024select}
Harry Walsh, Ben Saunders, and Richard Bowden.
\newblock Select and reorder: A novel approach for neural sign language production.
\newblock In \emph{Proceedings of the 2024 Joint International Conference on Computational Linguistics, Language Resources and Evaluation (LREC-COLING 2024)}, pages 14531--14542, 2024{\natexlab{b}}.

\bibitem[Walsh et~al.(2024{\natexlab{c}})Walsh, Saunders, and Bowden]{walsh2024sign}
Harry Walsh, Ben Saunders, and Richard Bowden.
\newblock Sign stitching: A novel approach to sign language production.
\newblock In \emph{The 35th British Machine Vision Conference (BMVC)}, 2024{\natexlab{c}}.

\bibitem[Xie et~al.(2022)Xie, Zhang, Li, Tang, Du, and Hu]{xie2022vector}
Pan Xie, Qipeng Zhang, Zexian Li, Hao Tang, Yao Du, and Xiaohui Hu.
\newblock Vector quantized diffusion model with codeunet for text-to-sign pose sequences generation.
\newblock \emph{arXiv preprint arXiv:2208.09141}, 2022.

\bibitem[Xie et~al.(2023)Xie, Peng, Du, and Zhang]{xie2023signlanguageproductionlatent}
Pan Xie, Taiyi Peng, Yao Du, and Qipeng Zhang.
\newblock Sign language production with latent motion transformer, 2023.

\bibitem[Xu et~al.(2022)Xu, Escalera, Pavão, Richard, Tu, Yao, Zhao, and Guyon]{Xu_2022}
Zhen Xu, Sergio Escalera, Adrien Pavão, Magali Richard, Wei-Wei Tu, Quanming Yao, Huan Zhao, and Isabelle Guyon.
\newblock Codabench: Flexible, easy-to-use, and reproducible meta-benchmark platform.
\newblock \emph{Patterns}, 3\penalty0 (7):\penalty0 100543, 2022.

\bibitem[Zelinka and Kanis(2020)]{9093516}
Jan Zelinka and Jakub Kanis.
\newblock Neural sign language synthesis: Words are our glosses.
\newblock In \emph{2020 IEEE Winter Conference on Applications of Computer Vision (WACV)}, pages 3384--3392, 2020.

\bibitem[Zuo et~al.(2024)Zuo, Wei, Chen, Mak, Yang, and Tong]{zuo2024simple}
Ronglai Zuo, Fangyun Wei, Zenggui Chen, Brian Mak, Jiaolong Yang, and Xin Tong.
\newblock A simple baseline for spoken language to sign language translation with 3d avatars.
\newblock In \emph{European Conference on Computer Vision}, pages 36--54. Springer, 2024.

\bibitem[Zwitserlood et~al.(2004)Zwitserlood, Verlinden, Ros, Van Der~Schoot, and Netherlands]{zwitserlood2004synthetic}
Inge Zwitserlood, Margriet Verlinden, Johan Ros, Sanny Van Der~Schoot, and T Netherlands.
\newblock Synthetic signing for the deaf: Esign.
\newblock In \emph{Proceedings of the conference and workshop on assistive technologies for vision and hearing impairment (CVHI)}, 2004.

\end{thebibliography}
}


\end{document}